\documentclass[runningheads]{llncs}

\usepackage{eccv}
\usepackage{algorithm}
\usepackage{algorithmic}
\usepackage{eccvabbrv}
\usepackage{graphicx}
\usepackage{booktabs}
\usepackage{cite}
\usepackage[accsupp]{axessibility}  
\usepackage[breaklinks,colorlinks,citecolor=eccvblue]{hyperref}
\usepackage{orcidlink}

\newcommand{\myparagraph}[1]{\vspace{0.1em}\noindent\textbf{#1}}

\begin{document}

\title{AC3S: Adaptive Conditioning for 3D-Aware Synthetic Data Generation} 

\author{Eric Ji\inst{1}\orcidlink{0009-0000-1811-2930} \and
Qiran Hu\inst{1}\orcidlink{0009-0007-1753-0189} \and Wufei Ma\inst{2}\orcidlink{0000-0002-4696-2833} \and Sarthak Jain\inst{1} \and \\ Yingying Li\inst{1}\orcidlink{0000-0002-1858-4257} \and Minh N. Do\inst{1, 3}\orcidlink{0000-0001-5132-4986} \and Yaoyao Liu\inst{1}\orcidlink{0000-0002-5316-3028}}

\authorrunning{E.~Ji et al.}

\institute{University of Illinois Urbana-Champaign \and
Johns Hopkins University \and College of Engineering and Computer Science, VinUniversity}

\maketitle

\begin{abstract}
Synthetic data generation has emerged as a powerful tool for improving data scalability in computer vision. Recent diffusion-based pipelines have demonstrated strong photorealism. However, how to enforce precise 3D structure and pose consistency in generated images remains challenging. Existing methods leverage visual prompts such as edge maps to guide diffusion models, but often suffer from over-conditioning artifacts that degrade image realism and limit dataset quality. In this paper, we present a diffusion-based image generation framework that enforces 3D structural alignment while preserving photorealism through adaptive conditioning. Our framework, Adaptive Conditioning for 3D-Aware Synthetic Data Generation (AC3S), introduces a self-supervised visual prompt modulator that dynamically adjusts the strength of ControlNet conditioning, preventing over-conditioning and enabling the diffusion model to retain its generative expressiveness. To further enhance diversity and semantic consistency, we develop a multi-agent vision language model framework that composes detailed and 3D-aware prompts aligned with the underlying geometric structure. Together, these components enable the scalable generation of high-quality synthetic datasets with accurate 2D and 3D annotations. Extensive experiments demonstrate that our method significantly improves image quality and downstream utility.\footnote[1]{Project page:  \href{https://ac3s.cvmlgroup.web.illinois.edu/}{\texttt{https://ac3s.cvmlgroup.web.illinois.edu/}}}
\end{abstract}

\section{Introduction}
\label{sec:intro}

Image generation has rapidly advanced in recent years due to the introduction of diffusion models\cite{ho2020denoising, song2021denoising, rombach2022high,cao2026freeorbit4d,Ma2024ImageNet3D,Liu2021Generating,duan2023prompt,cao2026redirect4dbench}. Large-scale pre-trained models can synthesize high-fidelity images spanning diverse domains\cite{podell2023sdxl, saharia2022photorealistic}. Beyond creative applications, these models have proven to be powerful tools for synthetic data generation, completely transforming the current landscape\cite{ma2023generating, he2022synthetic, nguyen2023dataset}.

Traditional synthetic datasets rely on physics-based rendering engines\cite{blender}, in which users directly control scene composition, lighting, and camera parameters. Building a photo-realistic scene is achievable, but not a trivial task, especially for large-scale datasets.

Diffusion models alternatively offer highly accessible photo-realistic capabilities through learning an underlying image data manifold. However, popular pre-trained text-to-image models lack reliable and precise 3D control. Recently, 3D-DST\cite{ma2023generating} bridged this gap with their image generation pipeline. Their method renders an image of a 3D model, extracts its edge map, and treats it as a visual prompt to condition a diffusion model through ControlNet\cite{zhang2023adding}. The objects in their generated images align in pose with the rendered model, allowing them to acquire 3D labels. While accurate poses are achieved, the framework sacrifices photorealism: backgrounds become blurred, object textures are simplified, and scenes become incoherent.

\begin{figure}[t!]
    \centering
    \includegraphics[width=0.9\linewidth]{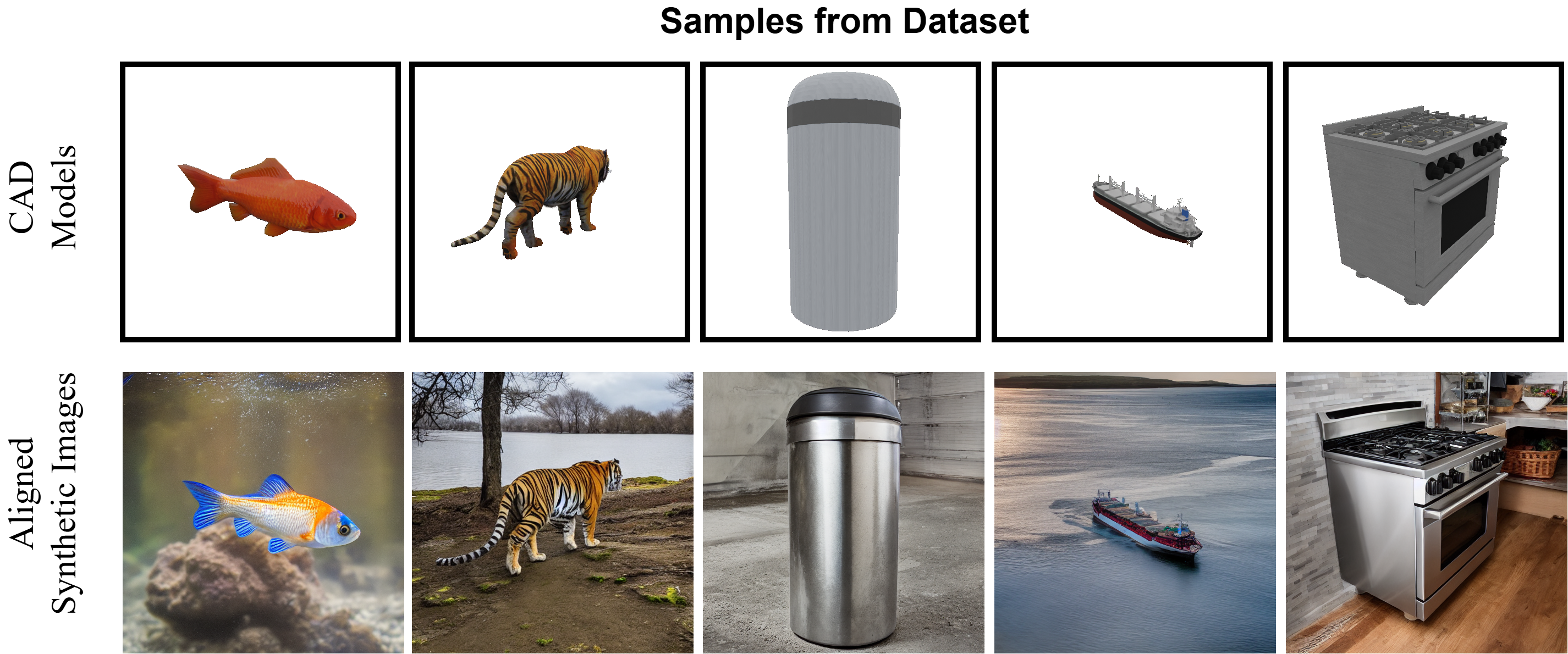}
    \caption{Illustrated are samples generated using our adaptive pipeline. The top row displays the CAD renders used to extract visual prompts and data labels. We combine the visual prompt with our 3D-aware text prompts to generate the synthetic images in the bottom row. Note that the subject in each synthetic image is precisely aligned with its corresponding render.}
    \label{fig:examples}
\end{figure}

In this paper, we propose a diffusion-based image generation framework that enforces 3D structural alignment while preserving the photorealistic capabilities of pre-trained diffusion models. Our key observation is that visual prompts derived from CAD renders are often sparse and overly restrictive when injected directly into diffusion models. To mitigate this issue, we introduce an adaptive visual prompt modulator that dynamically adjusts the strength of ControlNet conditioning during generation. The modulator predicts a conditioning scale from the feature magnitudes of the diffusion model and ControlNet, enabling the system to automatically balance structural alignment and generative expressiveness.

In addition to visual conditioning, generating diverse and semantically consistent prompts remains an important challenge in large-scale synthetic dataset creation. Template-based prompts often lack detail, while captions extracted from real images may conflict with the visual prompts used for geometric control. To address this issue, we design a multi-agent vision language model framework that composes detailed and 3D-aware prompts aligned with the underlying visual structure. Multiple specialized agents collaborate to identify the subject, infer pose information, and construct scene descriptions to generate both positive and negative prompts. This design ensures that textual guidance remains consistent with the visual prompts and encourages the diffusion model to generate realistic, diverse images.

Qualitatively, our pipeline's adaptive conditioning modules yield superior samples compared to the baseline. We validate such observations through a series of experiments analyzing generative quality and downstream utility. Specifically, we observe an improvement in FID score\cite{heusel2017gans} by 15.95 points over the baseline and provide up to a 7\% boost in accuracy when fine-tuning models with our synthetic data. Furthermore, we show that in the absence of real-world data, training strictly on our synthetic data can still produce functional models. 

Our contributions can be summarized as follows:
\begin{itemize}
    \item \textbf{Visual prompt modulator for ControlNet.} We introduce a simple yet effective self-supervised training strategy to train a visual prompt modulator. By automatically attenuating ControlNet outputs, we can prevent overconditioning on a per-sample basis, leading to higher quality images.
    \item \textbf{Multi-agent VLM prompt generator.} We propose a novel multi-agent framework to synthesize descriptive and diverse text prompts. By integrating 3D context, we can ensure alignment between text and visual prompts.
    \item \textbf{Full Synthetic Dataset.} Our full synthetic dataset consists of one million images evenly distributed among the 1,000 ImageNet\cite{deng2009imagenet} categories. We provide labels for classification, object pose estimation, and additional metadata. 
\end{itemize} 

\section{Related Work}

\myparagraph{Text-to-Image Diffusion Models.} Pre-trained text-to-image diffusion models excel in generating photo-realistic images from natural language\cite{podell2023sdxl, saharia2022photorealistic,Ho2020DiffusionModels,Singer2022Make,Villegas2022Phenaki,trabucco2023effective,Azizi2023Synthetic}. These models begin with sampling Gaussian noise, before iteratively passing through a denoising U-Net\cite{ronneberger2015u} until reaching the image manifold. Latent diffusion models\cite{rombach2022high} further optimize by operating in a compressed latent representation, addressing the high-dimensionality of images. Conditioning on text allows for guidance during generation towards desired results. Text is highly expressive, but it typically cannot provide fine-grained constraints. As a result, standard text-to-image models struggle with accurate geometric control, motivating alternative prompting techniques like ControlNet\cite{zhang2023adding}.

\vspace{0.2cm}

\myparagraph{ControlNet.} ControlNet\cite{zhang2023adding} extends the conditioning capabilities of diffusion models with visual prompts. Examples of visual prompts include edge maps, depth maps, and segmentation masks. Their method is inspired by side-tuning~\cite{zhang2020side}, an incremental learning~\cite{liu2020mnemonics,liu2021adaptive,liu2021rmm,liu2023online,luo2023class,liu2023continual,zhang2023continual,liu2024wakening,fischer2024inemo,duan2023prompt,zhu2025teachlmm,li2020online,li2019online,li2021online} technique that allows networks to adapt to new tasks by training a lightweight "side" network. For diffusion models, a copy of the U-Net encoder\cite{ronneberger2015u} acts as the "side" network. This ControlNet takes the visual prompt as input and learns to produce features to inject back into the decoder through skip connections. The injections are performed by element-wise addition
\begin{equation}
    \hat{f} = f + \mathcal{F}(I_{prompt};\theta)
\end{equation}
where $f$ denotes the intermediate U-Net features, and $\mathcal{F}(v;\theta)$ denotes the ControlNet outputs produced by visual prompt $v$ with parameters $\theta$. When training a ControlNet, the diffusion model's weights are frozen, allowing only the parameters $\theta$ to learn. In practice, we observe that features produced by pre-trained ControlNets are too strong and degrade photorealism, hence motivating our adaptive modulator.

\vspace{0.2cm}

\myparagraph{Vision Language Models (VLMs).} Vision language models (VLMs)~\cite{achiam2023gpt,bai2023qwen,zhu2025teachlmm,zhang2024vision,bordes2024introduction,zhu2024minigpt,zhou2022learning} have experienced widespread adoption with the introduction of GPT-4~\cite{achiam2023gpt} and QWEN~\cite{bai2023qwen}. Their ability to interpret both natural language and images makes them shine in descriptions and complex reasoning. VLMs generate their output by autoregressively generating one word at a time. Compared to prompting a single VLM, a multi-agent framework can decompose and more efficiently coordinate complex tasks. We assemble a team of agents to construct 3D-aware text prompts for image generation.

\vspace{0.2cm}

\myparagraph{3D-DST.} 3D-DST~\cite{ma2023generating} introduces a general framework for producing synthetic datasets with geometric control. Their method renders images of CAD models from diverse viewpoints to extract edge maps, which are used as visual prompts for ControlNet. Combined with text captions, their pipeline enables the generation of synthetic images with 3D annotations. While effective at enforcing geometric consistency, the lack of flexibility in their prompting leads to visible artifacts, the core problem our adaptive conditioning modules solve.

\section{Method}
Our method adaptively controls both visual and textual conditioning during image generation. By automatically tuning conditioning signals, our approach brings scalable solutions to limitations associated with prior pipelines. In particular, our proposed pipeline mitigates visual overconditioning, which manifests as low textures, and ensures consistency between visual and textual guidance. We begin by introducing the procedure for collecting visual prompts on a large-scale. Then, we discuss the motivation for the visual prompt modulator and the self-supervised training algorithm. Finally, we will lay out the design of our multi-agent VLM text prompt generator.

\begin{figure}[t!]
    \centering
    \includegraphics[width=0.9\linewidth]{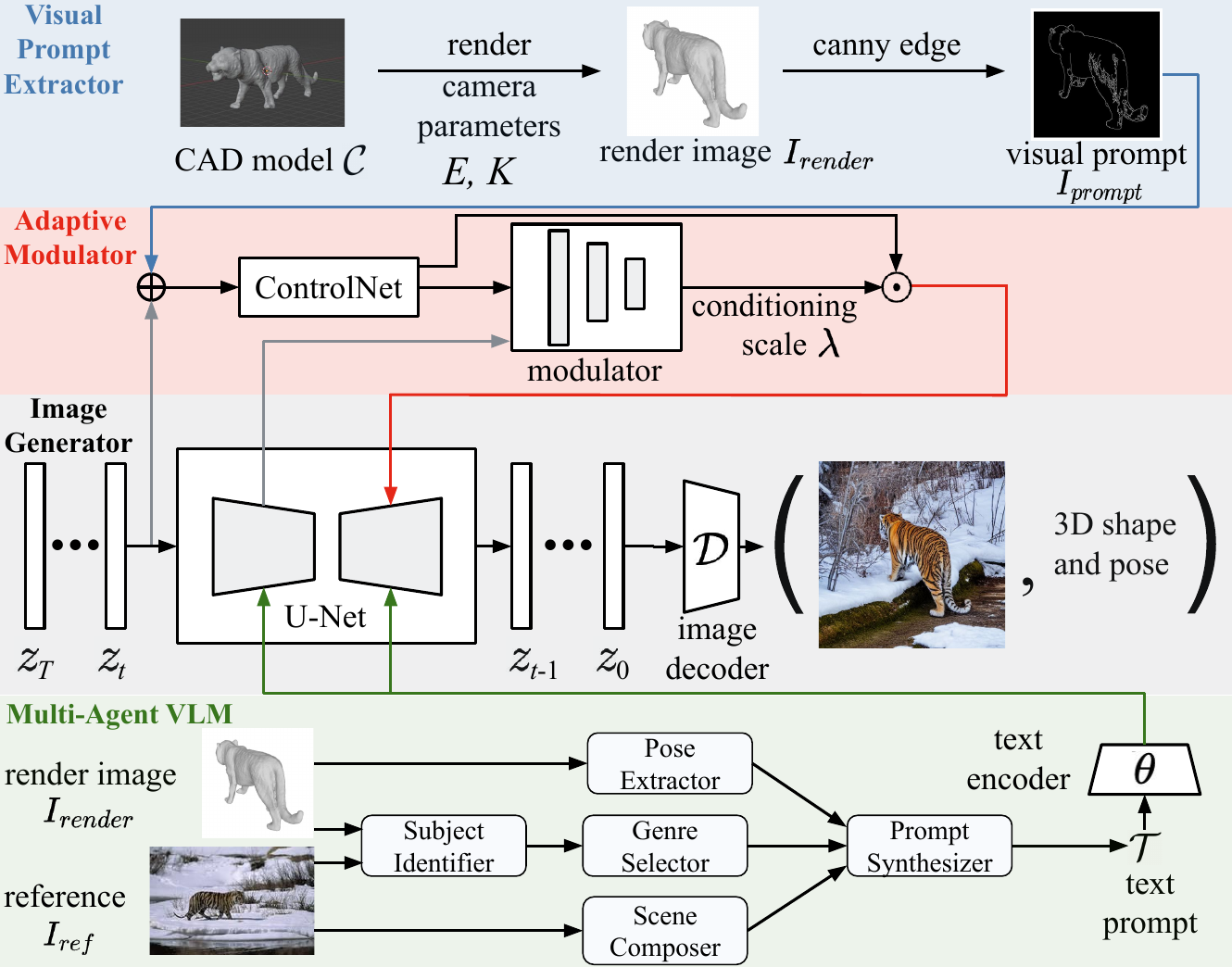}
    \caption{Our full generation pipeline is composed of four interconnected modules. (1) Visual Prompt Extractor: For a given 3D CAD model, we sample a set of camera parameters $E, K$ to render a 2D image $I_{render}$. The visual prompt $I_{prompt}$ is extracted from $I_{render}$ in the form of an edge map and serves as the input to ControlNet. (2) Adaptive Modulator: Instead of directly injecting ControlNet's output into the diffusion model, we treat it as input to our modulator alongside diffusion model features to predict the conditioning scale $\lambda$. Once attenuated by $\lambda$, ControlNet outputs are injected. (3) Multi-Agent VLM: The render $I_{render}$ and real image reference $I_{ref}$ are simultaneously processed by multiple agents to extract pose, color, and scene information for composing text prompt $\mathcal{T}$. (4) Image Generator: Our semantically consistent prompts are used to guide the generation of the photorealistic image.}
    \label{fig:full_pipeline}
\end{figure}

\subsection{Acquiring Visual Prompts Through Rendering}
Following 3D-DST\cite{ma2023generating}, we obtain visual prompts through graphics-based rendering. All 3D models are sourced from large-scale CAD repositories, including ShapeNet\cite{chang2015shapenet}, Objaverse\cite{deitke2023objaverse}, and OmniObject3D\cite{wu2023omniobject3d}, resulting in approximately 30 CAD models for each class category. For each model $C$, we randomly sample a camera viewpoint to render image $I_{render}$ using an off-the-shelf renderer $R$\cite{blender}.
\begin{equation}
    I_{render}= R(C, E, K)
\end{equation}
The camera's intrinsics $K$ are based on a perspective camera with a focal length of 35 mm. Camera extrinsics $E$ are sampled through varying distance, azimuth, and elevation. The distribution of viewpoints has defined constraints to avoid implausible perspectives that would not appear in real-world imagery. $I_{render}$ is not intended to be photorealistic, requiring no modifications to materials, lighting, or background. Instead, it is used as a source of geometric structure. We use a Canny edge detector\cite{canny2009computational} to extract an edge map for the visual prompt, 
\begin{equation}
    I_{prompt}= CannyEdge(I_{render})
\end{equation}
We explored alternative visual prompts, such as depth, but found that edge maps consistently yielded the highest quality samples.

\subsection{Adaptive Visual Prompt Modulator}
Most visual prompts are sparse in nature, particularly when computed from material-less and background-less renders. Feeding these prompts into ControlNet\cite{zhang2023adding} leads to overconditioning, manifesting as oversimplified textures and uninteresting scenes. Injecting unmodulated ControlNet outputs into our diffusion model heavily restricts the pre-trained generative prior that we want to leverage. 

In our study, we find that adding textual conditioning helps lower a diffusion model's denoising objective. Interestingly, this contrasts with injecting visual conditioning, which hurts the loss. This suggests ControlNet introduces harmful noise to the system. We find that attenuating ControlNet outputs can directly mitigate the damage. This appears as increased diversity of backgrounds, richer details, and an overall greater realism. 

We define this attenuation using a conditioning scale $\lambda$ where $f$ denotes the intermediate U-Net features of the diffusion model, and $\mathcal{F}(I_{prompt};\theta)$ denotes the ControlNet outputs conditioned on visual prompt $I_{prompt}$.

\begin{equation}
    \hat{f} = f + \lambda \cdot \mathcal{F}(I_{prompt};\theta)
\end{equation}

Manually tuning $\lambda$ can be effective for small batches, but scaling up for dataset generation would be infeasible. We find that the optimal $\lambda$ cannot generalize across class categories and even varies with different random seeds for the same $I_{prompt}$. However, we do note a correlation between $\lambda$, $f$, and $\mathcal{F}(I_{prompt};\theta)$. This motivated us to introduce our adaptive visual prompt modulator that predicts $\lambda$ dynamically based on $\lVert f\rVert$ and $\lVert \mathcal{F}(v;\theta)\rVert$ at the first timestep. Our implementation for the modulator is as a lightweight three-layer Multi-Layer Perceptron (MLP) as seen in Fig. \ref{fig:full_pipeline}.

Despite similar modulators\cite{lin2025tasr, yehezkel2025navigating} employing a denoising loss for supervision, we have shown it is problematic as the objective is trivially minimized with $\lambda=0$. To avoid a degenerate solution, we use self-supervision with pseudo-labels. Since $\lambda$ is continuous, we discretize into candidate bins. $\lambda$ values in the range $[0.3, 1.0]$ with increments of 0.1 are considered. We find that images generated with values below 0.3 almost never align with the visual prompt. When generating a sequence of images with $\lambda$ spanning the range, we observe that the desired 3D structure is suddenly enforced once $\lambda$ exceeds a threshold. Images in the sequence with $\lambda$ below this threshold behave as if barely affected by ControlNet. While pose exhibits a sudden transition, image quality gradually decreases as $\lambda$ increases. This phenomenon allows us to define the optimal $\lambda$ as the smallest $\lambda$ above the threshold.

We similarly define the optimal $\lambda$ as the point where the change in pose becomes distinguishable, which is analogous to the concept of "just noticeable difference" (JND). When provided with two images, JND refers to when a visible attribute changes just enough such that it is distinguishable by human observers\cite{yu2015just}. In our setting, the relevant attribute is pose, but other attributes can include shapes, expressions, or material properties.

\begin{figure}[t]
    \centering
    \includegraphics[width=1.0\linewidth]{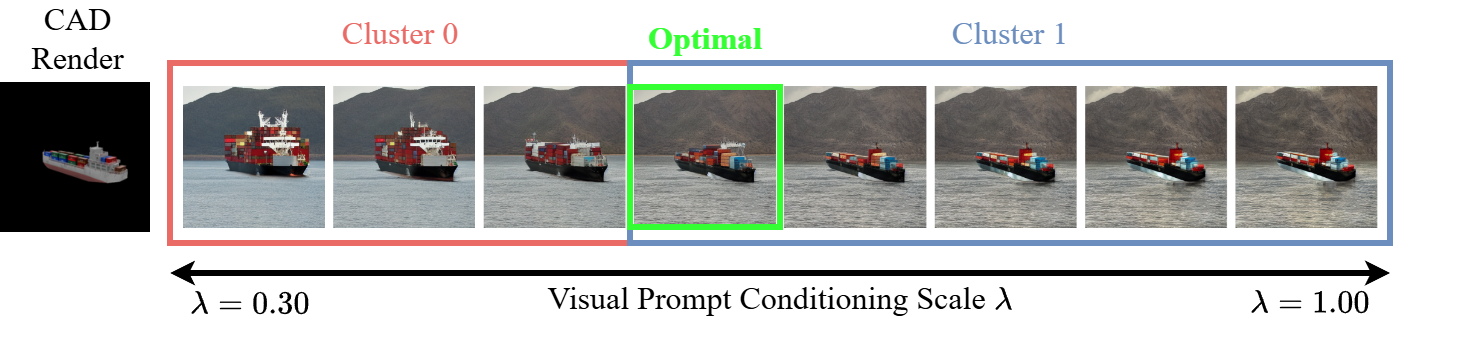}
    \caption{Performing clustering across a sequence of images generated with varying conditioning scales can be used to identify the optimal conditioning scale.}
    \label{fig:clustering}
\end{figure}

We utilize Algorithm \ref{alg:data_labeling} to acquire these pseudo-labels. For a render, we obtain the visual and text prompts to generate a sequence of images across a sweep of $\lambda$ values. Images in the sequence naturally observe two distinct distributions: (i) poorly aligned images at low $\lambda$, and (ii) aligned images with $\lambda$ above a threshold. This allows us to apply k-means clustering with $k=2$ on the pixel-level features to detect the JND. After clustering, we identify the cluster containing the image produced with $\lambda=1.0$ and designate it as the aligned cluster. The optimal $\lambda$, is then selected as the smallest $\lambda$ value within that cluster. Fig. \ref{fig:clustering} provides a visualization of this algorithm. We repeat this procedure across all class categories to obtain a set of pseudo labels, which are used to train the modulator without the need for any human annotations.

\begin{algorithm}[t]
\caption{Optimal Visual Conditioning Scale - Pseudo-Labeling}
\label{alg:data_labeling}
\begin{algorithmic}[1]
\STATE {\bfseries Input:} object renders $\{r_i\}_{i=1}^n$, candidate scales $\Lambda = \{0.30, 0.40, 0.50, \dots, 1.0\}$
\FOR{each render $r_i$}
    \STATE Extract edge map $v_i$ from $r_i$ as visual prompt
    \FOR{each $\lambda \in \Lambda$}
        \STATE Generate image $I_\lambda$ using ControlNet with scale $\lambda$ and prompt $v_i$
    \ENDFOR
    \STATE Cluster $\{I_\lambda : \lambda \in \Lambda\}$ using k-means ($k=2$)
    \STATE Let $C_{\lambda=1}$ be the cluster label of $I_{\lambda=1}$
    \STATE Optimal $\lambda^* = \min_{\lambda \in \Lambda} \text{ s.t. cluster}(I_\lambda) = C_{\lambda=1}$
\ENDFOR
\end{algorithmic}
\end{algorithm}

\subsection{Multi-Agent VLM Text Prompt Generator}
We introduce a multi-agent VLM system designed to synthesize text prompts that are semantically consistent with our visual prompts. This framework addresses several limitations associated with current approaches. Early methods~\cite{he2022synthetic} rely on fixed templates (e.g., "A high-quality image of a [CLASS]"), often lacking granular details and providing insufficient guidance for synthesizing complex scenes. More recent works mitigate this by leveraging VLMs as captioners to extract descriptions from large-scale image datasets \cite{ma2023generating, nguyen2023dataset}. While these captions are rich in detail, they frequently are not consistent with our visual prompts. For example, a sampled caption may describe a close-up scene containing multiple object instances, whereas our visual prompt depicts a holistic view of a singular instance. These conflicting text and visual prompts can produce degrading hallucinations. We orchestrate a team of specialized agents to analyze inputs, extract information, and compose quality prompts.

In our multi-agent system, each VLM acts as an independent expert defined by its role and domain expertise. Agents are profiled with detailed specifications, including input modalities, output schema, and behavioral constraints. We will highlight each agent and their objectives:

\myparagraph{Agent 1 (Subject Identifier)} takes the render $I_{render}$ and a real reference image $I_{ref}$ from ImageNet\cite{deng2009imagenet} to identify a subject label (1-3 words). Relying purely on basic ImageNet class categories instead of detailed labels can leave out meaningful context for downstream agents. By having access to both images, Agent 1 maximizes classification capabilities through cross-referencing before committing to a semantic anchor.

\myparagraph{Agent 2 (Genre Selector)} selects a photography genre from a pre-defined set (e.g., wildlife, macro, etc.) that would best align with the identified subject. Photography genres encode a prior on style and other conventions.

\myparagraph{Agent 3 (Pose Extractor)} takes the render $I_{render}$ and extracts the subject and camera pose. Since the edge map is extracted from the same render, using this information for our prompt will ensure alignment in perspective.

\myparagraph{Agent 4 (Scene Composer)} takes the reference image $I_{ref}$ to extract visual attributes for the subject (color, texture, etc.) and the background scenery. Having information from the reference allows prompts to capture details and diversity present in ImageNet.

\myparagraph{Agent 5 (Positive Prompt Synthesizer)} assembles the extracted information from agents 2-4 into a coherent positive prompt.

\myparagraph{Agent 6 (Protected Terms Identifier)} identifies keywords and phrases to exclude from the negative prompt. Synonyms from our positive prompt should be protected and avoided in the negative prompts to prevent inadvertently guiding the generation process away from the desired result.

\myparagraph{Agent 7 (Negative Prompt Synthesizer)} assembles a negative prompt by identifying common negative terms applicable to our subject, while ensuring they do not include protected terms produced by Agent 6.

\subsection{Prompt Engineering for Agents}
We build each agent using well-established prompt-engineering principles like chain-of-thought prompting\cite{wei2022chain} and structured output generation\cite{white2023prompt}. Each agent operates according to a structured prompt template consisting of seven standardized sections: Role, Goal, Context, Constraints, Contrastive Examples, Output Format, and Criteria\cite{robino2025conversationroutinespromptengineering}. Table \ref{tab:agent_prompt} outlines the purpose of each section.

\begin{table}[t]
  \centering
  \caption{To ensure each agent specializes in their designated tasks, we provide structured prompts following the format outlined in this table.} 
  \label{tab:agent_prompt}
  \begin{tabular}{@{}lll@{}}
    \toprule
    Section & Purpose\\
    \midrule
    Role & Initializes an expert persona aligned with the agent's specialized task.\\
    Goal & Provides a single-sentence objective that defines success criteria.\\
    Context & Describes input modalities and their semantic content.\\
    Constraints & Enumerates explicit behavioral boundaries to eliminate uncertainty.\\
    Examples & Provides paired positive and negative demonstrations with rationales.\\
    Format & Specifies the expected structure of outputs for reliable parsing.\\
    Criteria & Defines acceptance criteria for self-verification.\\
  \bottomrule
  \end{tabular}
\end{table}

The multi-agent workflow is implemented using LangGraph's StateGraph abstractions, which provide directed acyclic graph execution with type state management. We employ Qwen3-VL-23B-Instruct\cite{bai2023qwen} with 4-bit quantization. Agents communicate through a shared typed state dictionary following the shared message pool pattern introduced by MetaGPT\cite{hong2023metagpt}. The communication structure combines sequential and parallel workflows as certain agents require dependencies from upstream agents. The workflow for producing the positive prompt is visualized in Fig. \ref{fig:full_pipeline}.

\subsection{Dataset Filtering}
Filtering out poorly generated samples from a dataset significantly boosts performance for downstream tasks. 3D-DST\cite{ma2023generating} trains a pose estimator with k-fold cross-validation to assign a loss-based score for each image. For smaller datasets, it may be difficult for the pose estimator to converge and produce a meaningful score. It is also extremely computationally expensive for large datasets, as it requires a designated model for each class category. We instead leverage DreamSim\cite{fu2023dreamsim}, an ensemble of foundation models fine-tuned to capture a holistic perceptual metric. Unlike traditional metrics, which target low-level features, DreamSim score primarily captures object pose and semantic information. To evaluate generated samples, we compute the perceptual similarity between each synthetic image and its corresponding render image $I_{render}$. We ensure the metric remains invariant to color by applying a grayscale transformation to both images before scoring. We find that using the DreamSim score as a ranking mechanism can efficiently allow us to identify the strongest samples for our dataset.

\section{Experiments}
To evaluate the effectiveness of our adaptive prompting techniques, we generate a dataset using the proposed pipeline and conduct several comprehensive experiments. We first conduct an ablation study analyzing the impact of individual modules on image quality. Beyond a qualitative assessment, we validate the utility of our dataset through studying its impact on both 2D and 3D computer vision tasks. We benchmark our methods against multiple existing synthetic datasets.

\subsection{Image Quality Ablation}
\begin{figure}[!t]
    \centering
    \includegraphics[width=0.9\linewidth]{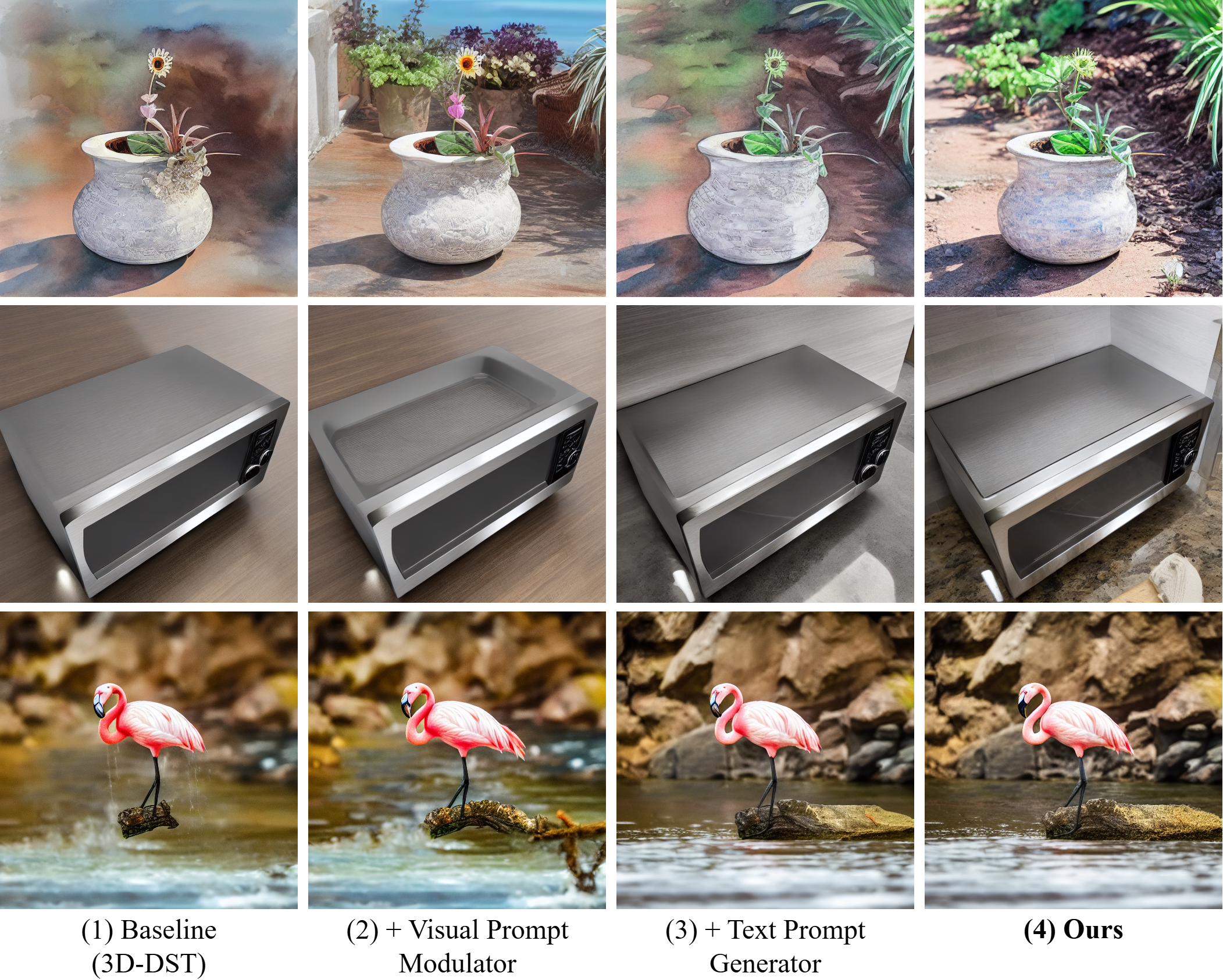}
    \caption{Visualizations of various classes from the ablation study. For the flamingo in the last row, our method (4) uses both the visual prompt modulator and text prompt generator simultaneously to produce the most realistic and detailed image. Compared to (3), introducing the visual prompt modulator in (4) brings out greater texture and details in both the bird itself and the rocky background. Simultaneously, using the prompts from our text prompt generator in (3) results in more realistic colors and plausible scenery compared to (1).}
    \label{figure:ablation}
\end{figure}

We conduct an ablation study for each module of our method to demonstrate the impact on image quality and diversity. We quantify quality and diversity through the Frechet Inception Distance (FID) score\cite{heusel2017gans}, which measures the distance between the distribution of our generated images and real images. In this experiment, we compare 1,000 synthetic images from each method to 1,000 images from ImageNet\cite{deng2009imagenet} for 50 classes. The results are outlined in Table~\ref{table:ablation}. The baseline method is 3D-DST\cite{ma2023generating}, to which we incorporate the adaptive visual prompt modulator and multi-agent VLM text prompt generator to produce our method. When generating the data across methods, we manually fixed the seeds to ensure fair comparisons. Analyzing the results, we can observe that both the visual prompt modulator and text prompt generator can improve the overall FID score by 9.66 and 8.26 points, respectively. When working in conjunction for our full pipeline, we observe the greatest overall improvement of 15.95 points. We can also note that our method particularly boosts image quality and diversity for animal classes with a 23.8 point improvement. \Cref{figure:ablation} illustrates example images from this ablation study.

\subsection{Synthetic Datasets}
In our model training experiments, we use AC3S-50, a 50-class dataset generated with our adaptive conditioning modules. The distribution of classes contains a mixture of animals and objects following the ImageNet~\cite{deng2009imagenet} distribution. For our diffusion model, we use Stable Diffusion v1.5 and sample 20-timestep trajectories. We benchmark this dataset against two existing synthetic datasets, Text2Img~~\cite{ma2023generating}, which we generate following their implementation and codebase. For each synthetic dataset, we have 1,000 images per class, equating to a total of 50,000 images each.

\begin{table}[!t]
\centering
\caption{Results from our ablation study analyzing the impact of each adaptive conditioning module.}
\label{table:ablation}
\begin{tabular}{@{}c@{}c@{}c|ccc@{}}
\toprule
& & & \multicolumn{3}{c}{FID Score ($\downarrow$)} \\
\cmidrule(l){4-6}
Method & Modulator & VLM Prompts & Animals & Objects & Overall \\
\midrule
Baseline (3D-DST\cite{ma2023generating}) & & & 72.67 & 90.38 & 87.19 \\
+ Visual Prompt Modulator & \checkmark &            & 60.93 & 81.18 & 77.53 \\
+ Text Prompt Generator &            & \checkmark & 54.68 & 84.23 & 78.93 \\
\textbf{Ours} & \checkmark & \checkmark & \textbf{48.87} & \textbf{76.15} & \textbf{71.24} \\
\bottomrule
\end{tabular}
\end{table}

\subsection{Synthetic Pre-training for Classification Implementation}
We evaluate the utility of synthetic data for classification across three methods. The baseline method is trained using only our real data from ImageNet\cite{deng2009imagenet}. The remaining three methods incorporate synthetic data during pre-training.

The most common case of synthetic data is to pre-train a model before fine-tuning it on the target domain with real data. In our classification experiments, we use ResNet-18\cite{he2016deep}, Swin-S\cite{liu2021swin}, and ConvNeXt-S\cite{liu2022convnet} to evaluate our dataset across convolutional and transformer-based backbones. All models are trained completely from scratch with default PyTorch\cite{paszke2017automatic} weight initialization. For ResNet-18, we fine-tune for 50 epochs on each synthetic dataset, which was sufficient for model convergence. We select the best pre-trained weights from the epoch with the lowest validation loss. We follow the same pre-training procedure for Swin-S and ConvNeXt-S, which required 150 epochs and 200 epochs, respectively. Following pre-training, each model is fine-tuned on ImageNet\cite{deng2009imagenet} until convergence, with the final model chosen from the epoch with the lowest validation loss. ResNet-18 and Swin-S each required 150 epochs, while ConvNeXt-S needed 500 epochs to converge. 

\subsection{Synthetic Pre-training for Classification Results}
We report all results as top-1 accuracies evaluated on a held-out test split of ImageNet\cite{deng2009imagenet}. Table~\ref{table:classification} summarizes the results for pre-training on synthetic data before fine-tuning with real data. We observe that across all three architectures, using our data for pre-training yields the greatest overall improvement. ConvNeXt-S\cite{liu2022convnet} and Swin-S\cite{liu2021swin} both saw a significant increase of ~2.6\%, and ResNet-18\cite{he2016deep} gained 0.99\%. 

\begin{table}[t]
\centering
\caption{Reported top-1 classification accuracies on ImageNet\cite{deng2009imagenet}. Each method, except for Baseline, was pre-trained on synthetic data before being fine-tuned to ImageNet.}
\label{table:classification}
\scriptsize
\begin{tabular}{@{}l ccc ccc ccc@{}} 
\toprule
& \multicolumn{3}{c}{Animals} & \multicolumn{3}{c}{Objects} & \multicolumn{3}{c}{Overall} \\ 
\cmidrule(lr){2-4} \cmidrule(lr){5-7} \cmidrule(l){8-10}
Method & ResNet & ConvNeXt & Swin & ResNet & ConvNeXt & Swin & ResNet & ConvNeXt & Swin \\
\midrule
Baseline    & 91.30 & 94.03 & 96.16 & 73.08 & 77.05 & 78.37 & 76.34 & 80.12 & 81.59 \\
Text2Img\cite{he2022synthetic} & 91.89 & 93.52 & 96.33 & 73.80 & 76.05 & 78.05 & 77.08 & 79.21 & 81.36 \\
3D-DST\cite{ma2023generating}   & 91.98 & \textbf{95.39} & 96.25 & 73.63 & 78.90 & 80.28 & 76.96 & 81.89 & 83.17 \\
\textbf{AC3S(Ours)} & \textbf{92.66} & 95.14 & \textbf{96.76} & \textbf{73.93} & \textbf{80.01} & \textbf{81.41} & \textbf{77.33} & \textbf{82.75} & \textbf{84.19} \\
\bottomrule
\end{tabular}
\end{table}

\subsection{Synthetic Pre-training for Pose Estimation Implementation}
We perform a similar pre-training procedure for object pose estimation to validate the accuracy of our 3D annotations. Instead of ImageNet\cite{deng2009imagenet}, we fine-tune and evaluate with PASCAL3D+\cite{xiang2014beyond}, which consists of approximately 30,000 images spread across 10 class categories. We selected 3 class categories that intersected with our AC3S-50 dataset: sofa, dining table, and car. For each class category and method, we trained a designated pose estimator with ResNet-18\cite{he2016deep} backbone. Pre-training required 20 epochs for model convergence, and fine-tuning to PASCAL3D+ required 10.

\subsection{Synthetic Pre-training for Pose Estimation Experiment Results}
We report all results as threshold accuracies in Table~\ref{table:pose_estimation} based on the geodesic distance between the predicted and ground-truth rotation matrix. Across all three class categories, our method consistently yielded greater improvement over 3D-DST\cite{ma2023generating}. On average, we observe a 1.23\% gain for $\pi/6$ threshold and 2.17\% gain for $\pi/18$. We find the biggest improvement over the baseline to be sofa for $\pi/18$ with a jump of 7.39\% by pre-training with AC3S. In Table~\ref{table:pose_estimation_synth_only}, without any fine-tuning, the gap performance between 3D-DST and AC3S is immense, implying that AC3S better resembles the distribution of PASCAL3D+.

\begin{table}[t]
\centering
\caption{Reported accuracies measure the geodesic distance between predicted and ground-truth rotation matrices with a $\pi/6$ and $\pi/18$ error threshold on PASCAL3D+\cite{xiang2014beyond}. Each method, except for Baseline, was pre-trained on synthetic data before being fine-tuned to PASCAL3D+.}
\label{table:pose_estimation}
\begin{tabular}{@{}l cc cc cc@{}}
\toprule
& \multicolumn{2}{c}{Sofa} & \multicolumn{2}{c}{Dining Table} & \multicolumn{2}{c}{Car}\\
\cmidrule(lr){2-3} \cmidrule(lr){4-5} \cmidrule(l){6-7}
Method & Acc@$\pi$/6 & Acc@$\pi$/18 & Acc@$\pi$/6 & Acc@$\pi$/18 & Acc@$\pi$/6 & Acc@$\pi$/18 \\
\midrule
ResNet (Baseline)   & 80.79 & 34.98 & 79.86 & 49.83 & 87.60 & 66.19 \\
w/ 3D-DST\cite{ma2023generating}   & 86.70 & 37.83 & 80.20 & 51.88 & 88.51 & 67.62 \\
\textbf{w/ AC3S (Ours)} & \textbf{88.18} & \textbf{38.92} & \textbf{81.91} & \textbf{53.92} & \textbf{89.03} & \textbf{71.02} \\
\bottomrule
\end{tabular}
\end{table}

\begin{table}[t]
\centering
\caption{Reported accuracies measure the geodesic distance between predicted and ground-truth rotation matrices with a $\pi/6$ and $\pi/18$ error threshold on PASCAL3D+\cite{xiang2014beyond}. Each method was only trained on synthetic data.}
\label{table:pose_estimation_synth_only}
\begin{tabular}{@{}l cc cc cc@{}}
\toprule
& \multicolumn{2}{c}{Sofa} & \multicolumn{2}{c}{Dining Table} & \multicolumn{2}{c}{Car}\\
\cmidrule(lr){2-3} \cmidrule(lr){4-5} \cmidrule(l){6-7}
Method & Acc@$\pi$/6 & Acc@$\pi$/18 & Acc@$\pi$/6 & Acc@$\pi$/18 & Acc@$\pi$/6 & Acc@$\pi$/18 \\
\midrule
w/ 3D-DST\cite{ma2023generating}   & 10.84 & 0.99 & 14.68 & 2.05 & 7.05 & 0.39 \\
\textbf{w/ AC3S (Ours)} & \textbf{34.48} & \textbf{10.34} & \textbf{33.79} & \textbf{5.46} & \textbf{47.91} & \textbf{17.49} \\
\bottomrule
\end{tabular}
\end{table}

\section{Conclusion}
In this paper, we present AC3S, a 3D-aware image generation pipeline centered on adaptive conditioning. Our self-supervised visual prompt modulator adaptively attenuates ControlNet\cite{zhang2023adding} outputs, effectively preventing over-conditioning. This allows us to strike a balance between geometric alignment and generative expressiveness. Simultaneously, our multi-agent VLM system establishes a novel approach to producing text prompts that remain consistent with our visual prompt. Experimentation confirms that when these two adaptive conditioning modules are integrated together, we not only get a significant boost in image quality, but also increased utility for downstream 2D and 3D vision tasks. We believe AC3S provides many meaningful insights into scaling and advancing synthetic image generation.\\

\noindent\textbf{Acknowledgements.} This research is supported by the National Artificial Intelligence Research Resource Pilot Awards NAIRR250199, NAIRR260019, and NAIRR260077, the AMD University Program’s AI \& HPC Cluster, NVIDIA Academic Grant Program, and Lambda's Research Grant. It is also supported in part by the Center for Networked Intelligent Components and Environments, a partnership between the University of Illinois Urbana-Champaign and Foxconn Interconnect Technology. Computational resources are also provided by Delta and DeltaAI at the National Center for Supercomputing Applications through ACCESS allocations CIS250012, CIS250816, and CIS251188. 

\bibliographystyle{splncs04}
\bibliography{main}
\end{document}